\documentclass{article}
\PassOptionsToPackage{numbers, compress}{natbib}

\usepackage[preprint]{arxiv}

\usepackage{amsmath,amsfonts,bm}

\def\eqref#1{equation~\ref{#1}}
\def\1{\bm{1}}

\DeclareMathAlphabet{\mathsfit}{\encodingdefault}{\sfdefault}{m}{sl}
\SetMathAlphabet{\mathsfit}{bold}{\encodingdefault}{\sfdefault}{bx}{n}

\usepackage{url}

\usepackage{xspace}
\usepackage{dsfont}
\usepackage{afterpage}

\usepackage{enumitem}
\usepackage{booktabs}
\usepackage{arydshln}
\usepackage{caption} % captionof for minipage
\usepackage{multirow} % multirow
\usepackage{enumitem}
\usepackage{colortbl} % rowcolor
\usepackage{bbm}
\usepackage{algorithm}
\usepackage{algpseudocode}                  % algorithm formatting
\usepackage{setspace}                       % pleasant display of algorithm.
\usepackage{lipsum} % placeholder
\usepackage{subcaption} % subtable
\usepackage{xcolor}

\usepackage{wrapfig}
\usepackage{pifont}

\usepackage{mathtools} % for coloneqq
\usepackage{bm} % bm
\usepackage{soul} % highlight text
\usepackage{nicefrac}
\usepackage{csquotes}

\usepackage{duckuments}

\makeatletter
\newcommand*{\myfnsymbol}[1]{%
  \ensuremath{%
    \ifcase#1\or
      \text{\ding{41}}\or
      1\or
      2\or
      3\or
      4\or
      5\or
      6\or
      7\else
      8\fi
  }%
}
\def\@fnsymbol#1{\myfnsymbol{#1}}
\def\thempfootnote{\myfnsymbol{\c@mpfootnote}}
\makeatother

\newcommand{\sname}{D-OPSD\xspace}

\definecolor{cornellred}{rgb}{0.7, 0.11, 0.11}
\definecolor{cadmiumgreen}{rgb}{0.0, 0.42, 0.24}
\definecolor{aliceblue}{rgb}{0.91, 0.94, 0.97}
\definecolor{darkblue}{rgb}{0.83, 0.89, 0.97}
\definecolor{Red7}{rgb}{0.941, 0.243, 0.243}
\definecolor{Green7}{RGB}{55, 178, 77}
\definecolor{Blue9}{rgb}{0.098,0.3,0.9}

\usepackage{pifont}% http://ctan.org/pkg/pifont

\sethlcolor{aliceblue}

\definecolor{citecolor}{HTML}{2980b9}
\definecolor{linkcolor}{HTML}{c0392b}
\definecolor{urlcolor}{RGB}{157,49, 251}

\usepackage[hidelinks,breaklinks=true,colorlinks,citecolor=citecolor,linkcolor=linkcolor,urlcolor=urlcolor]{hyperref}

\title{

\sname : On-Policy Self-Distillation for Continuously Tuning Step-Distilled Diffusion Models

}

\author{ 
\vspace{-39pt}\\
\hspace{-6mm} \textbf{Dengyang Jiang}$^{1,2}$\hspace{2mm}
\textbf{Xin Jin}$^{2}$\hspace{2mm} 
\textbf{Dongyang Liu}$^{4,2}$\hspace{2mm} 
\textbf{Zanyi Wang}$^{3}$\hspace{2mm} 
\textbf{Mingzhe Zheng}$^1$ \hspace{2mm}
\textbf{Ruoyi Du}$^2$ \\ [0.5mm]
\hspace{-6mm} \textbf{Xiangpeng Yang}$^2$\hspace{2mm}
\textbf{Qilong Wu}$^2$ \hspace{2mm}
\textbf{Zhen Li}$^2$\hspace{2mm}
\textbf{Peng Gao}$^2$\thanks{Corresponding authors} \hspace{2mm}
\textbf{Harry Yang}$^1$\textsuperscript{\ding{41}} \hspace{2mm} 
\textbf{Steven C.H. Hoi}$^2$\\ [1mm]
\hspace{-8mm} $^1$The Hong Kong University of Science and Technology \hspace{1mm} $^2$Z-Image Team, Alibaba Group \\[0.5mm]
\hspace{-6mm}$^3$University of California, San Diego  \hspace{3mm} $^4$The Chinese University of Hong Kong\\[1mm]
 \hspace{-4mm} \url{https://vvvvvjdy.github.io/d-opsd}
 }

\begin{document}

\maketitle

\begin{abstract}
The landscape of high-performance image generation models is currently shifting from the inefficient multi-step ones to the efficient few-step counterparts (e.g, Z-Image-Turbo and FLUX.2-klein). However, these models present significant challenges for direct continuous supervised fine-tuning. For example, applying the commonly used fine-tuning technique would compromise their inherent few-step inference capability. To address this, we propose \textbf{\sname}, a novel training paradigm for step-distilled diffusion models that enables on-policy learning during supervised fine-tuning. We first find that the modern diffusion models, where the LLM/VLM serves as the encoder, can inherit its encoder's in-context capabilities. This enables us to formulate the training as an on-policy self-distillation process. Specifically, during training, we make the model act as both the teacher and the student with different contexts, where the student is conditioned only on the text feature, while the teacher is conditioned on the multimodal feature of both the text prompt and the target image. Training minimizes the two predicted distributions over the student's own roll-outs. By optimizing on the model's own trajectory and under its own supervision, \sname enables the model to learn new concepts, styles, etc., without sacrificing the original few-step capacity.
\end{abstract}

\section{Introduction}
\label{sec:intro}
Recent years have seen significant progress in text-to-image (T2I) generation, with models advancing from synthesizing rudimentary textures to producing images that exhibit strong adherence to semantic descriptions~\cite{z-image,flux-1,flux-2,sd3,qwenimage,hunyuanimage,seedream4,nanopro,gptimage-1}. However, the sampling process typically requires numerous iterative denoising steps~\cite{ddim,ddpm,flow-matching}, leading to substantial latency and computational cost in practice. To address this, researchers have developed various step-distillation techniques~\cite{lcm,dmd,diff-instruct,dmd2,piflow} that substantially reduce the number of function evaluations (NFEs). Furthermore, recent advances in distillation methodology~\cite{ddmd,dmdr,dmd2,twinflow,tdmr1} have enabled state-of-the-art open-source few-step diffusion models to surpass their multi-step predecessors not only in sampling efficiency but also in generated image quality. As a result, such few-step models are increasingly adopted in practical production settings.

Despite these advances, how to continually finetune these step-distilled diffusion models remains unclear. A straightforward solution is to apply the standard supervised fine-tuning objective used in the multi-step counterpart~\cite{flow-matching,rectified-flow}, i.e., feeding a noised target image into the model and supervising it with the corresponding flow-matching target.\footnote{In this paper, we mainly discuss flow-matching models, as they are currently the default choice in the field.} However, this training signal is defined on externally induced states of the target image that belong to an offline data distribution, rather than on the states actually visited by the model's own few-step sampler. For step-distilled models, whose generation quality relies on a small number of carefully distilled denoising updates, such a mismatch can easily perturb the learned few-step dynamics and degrade inference quality. This effect is also borne out empirically: across our experiments, and echoed by community reports, standard SFT often compromises the model’s original distilled few-step ability to generate high-quality images. Online Reinforcement learning (RL), in contrast, would not impair the few-step capabilities when used as the training algorithm for the model~\cite{dmdr,diffinstruct++}. This is because it optimizes the model on samples generated by the current model and derives the learning signal on the same sampled trajectory. However, it requires a well-designed reward function~\cite{refl,flowgrpo,dancegrpo,diffusionnft,hpsv2,clipscore,pickscore},
which is not feasible for most secondary developers in the community, as they usually have only the image-text pairs to customize concepts. 

 Thus, we assume that a suitable continuous-tuning strategy should be a combination of the two: it should update the model on its own roll-outs, and it should incorporate supervision from paired image-text data on those same visited states. A natural candidate is on-policy self-distillation (OPSD), which has recently been studied in autoregressive large language models (LLMs)~\cite{sdft,sdrlvr,opsd,sd-zero,pid}. OPSD retains the appeal of on-policy learning while avoiding explicit reward design: the model samples from its current policy as a student, while a stronger teacher distribution is obtained by conditioning the same model on richer in-context information~\cite{chain-of-thought,gpt3}. This perspective is particularly appealing in our setting because the target image in each training pair naturally provides the supervision. However, directly transferring the idea of OPSD to diffusion models is nontrivial. In text generation with LLMs, the  context can simply be appended to the input sequence. In diffusion models, by contrast, directly feeding the target image into the denoising process would alter the trajectory itself, reducing the formulation back to the standard off-policy SFT regime. The key challenge is therefore: \textit{how can target-image information be introduced as stronger context while keeping the student's few-step roll-out unchanged?}

\begin{figure}[t]
    \centering
    \vspace{-1.5em}
    \includegraphics[width=1\linewidth]{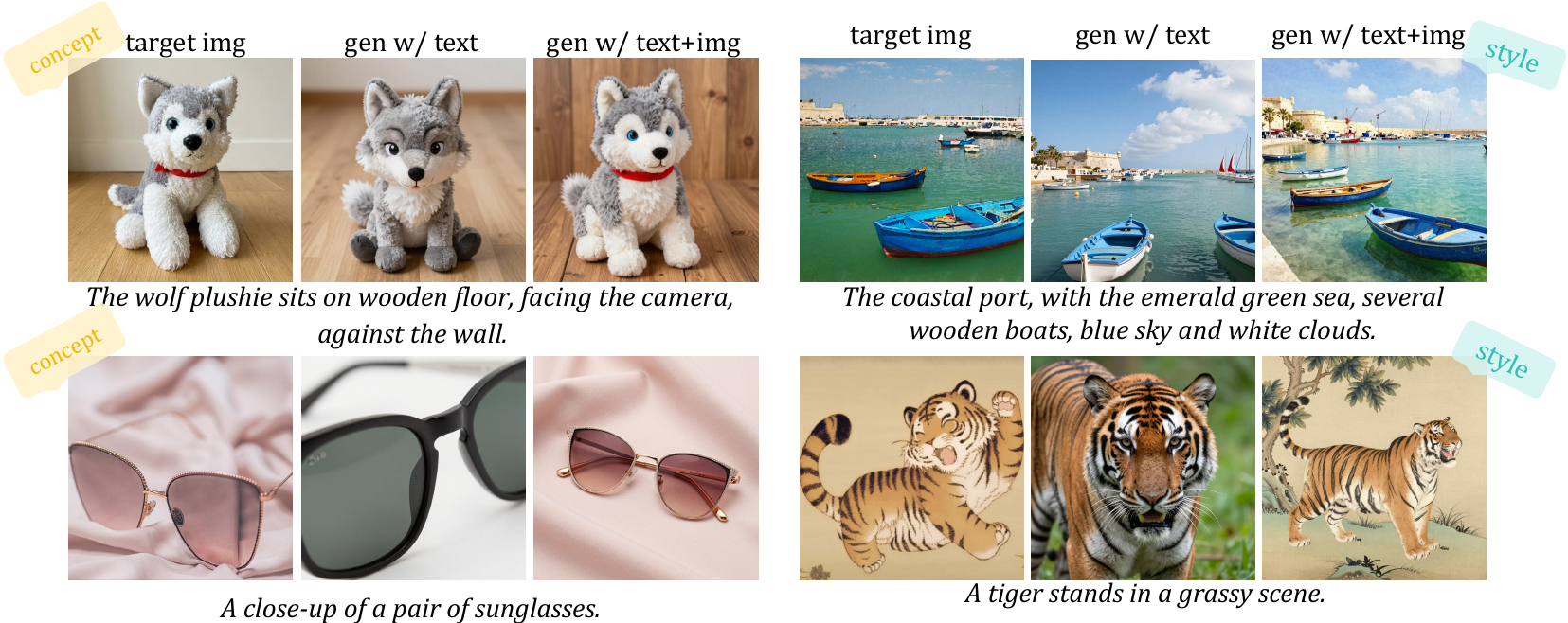}  
    \caption{We empirically  investigate the visual appearance of generated images when conditioned on only the text prompt feature or the multimodal feature of target image and the text prompt using Z-Image Turbo~\cite{z-image} with 8 steps. Investigation of  FLUX.2-klein~\cite{flux-2} is provided in Appendix~\ref{appen:flux_zeroshot}.}
    \vspace{-1em}
    \label{fig:motivation}
\end{figure}

We address this challenge by proposing \textbf{\sname}. Unlike earlier T2I diffusion models that used T5~\cite{t5} or CLIP~\cite{clip} as the encoder~\cite{sdxl,flux-1,sd3}, current state-of-the-art diffusion models increasingly adopt LLM/VLM backbones~\cite{qwen2vl,qwen3} as their encoders~\cite{flux-2,z-image,qwenimage}. This raises a natural question: can the subsequent diffusion model inherit the encoder's in-context capability? As shown in Figure~\ref{fig:motivation}, we find that the answer is yes. When replacing text-only features with multimodal features extracted from both the text prompt and the target image, the diffusion model can already generate variations that preserve the target concept or style (see gen w/text+img), even without any additional training. This emergent behavior enables us to instantiate OPSD in diffusion models. Specifically, during training, we assign the same model two roles: a student conditioned only on the text feature, and a teacher conditioned on the multimodal feature of the text prompt and the target image. We then distill the teacher's predictions into the student along the student's own roll-outs, yielding a one-stage on-policy framework that injects target-image information without requiring external modules or reward design.

We evaluate \sname in the settings of both LoRA training on small customized dataset and full fine-tuning on larger dataset, the results show that our  method enables the model to acquire new knowledge (e.g, specific concept, style) from the target image-text pair while preserving its original few-step inference capability. Furthermore, rather than learning via overfitting to the training pair, the acquired knowledge with our method demonstrates strong generalization across unseen prompts (e.g, generating training concepts in different scenarios). These results suggest promising prospects for the continual learning of step-distilled diffusion models.

In summary, our main contributions are as follows:
\vspace{-0.07in}
\begin{itemize}[leftmargin=*,itemsep=0mm]	
		\item We identify an emergent property of modern text-to-image diffusion models with LLM/VLM encoders and utilize this property to the continuous tuning of the step-distilled diffusion model.
		\item We propose \sname, a novel diffusion model on-policy self-distillation framework. By assigning the same model two roles with different contexts, \sname enables supervised tuning on the student's own roll-outs without requiring any external reward function or extra modules.
		\item We validate \sname in different settings. The results show that our method enables the model to learn new concepts, styles, and domain preferences while preserving its original few-step inference capability and previous knowledge.
\end{itemize}

\section{Method}
\subsection{Background}

In this study, our goal is to continually tune a step-distilled diffusion model on  image-text pairs while preserving its original few-step inference capability. As discussed in Section~\ref{sec:intro}, this is difficult for conventional fine-tuning. Vanilla SFT optimizes the model on noised target images rather than on the states visited by its own sampler, and the supervision is provided by an external target velocity that is unavailable at inference time~\cite{flow-matching,rectified-flow}. Such a train test mismatch may make the model acquire new concepts or styles at the cost of distorting the previously distilled few-step generation distribution (distribution shift). Online RL-style methods are more compatible with this setting because they optimize the model on its own roll-outs and derive supervision from the same on-policy samples~\cite{flowgrpo,dancegrpo,refl}, but they rely on carefully designed reward functions or preference signals~\cite{hpsv2,pickscore}, which are typically unavailable in practical customization scenarios. We address this gap by constructing an on-policy self-distillation framework for diffusion models, which uses only paired image-text data and does not require any external reward.

\subsection{\sname}

\paragraph{OPSD in LLMs and our solution for implication in diffusion models.}
On-policy self-distillation (OPSD) is first proposed in language models with a simple idea: the same model can act as both a student and a teacher under different contexts. Given an input query $q$, let $r$ denote additional in-context information, such as demonstrations, intermediate reasoning, or the ground-truth response~\cite{opsd,sdft,sd-zero,opsdc,pid}. The student predicts under the weaker context $q$, while the teacher predicts under the stronger context $(q,r)$. Let $\pi_\theta(\cdot \mid q)$ denote the student distribution and $\pi_{\bar{\theta}}(\cdot \mid q,r)$ the teacher distribution. OPSD optimizes the student on its own sampled outputs $\hat{o} \sim \pi_\theta(\cdot \mid q)$,
and minimizes a divergence between the teacher and student predictions on that on-policy sample:
\begin{equation}
\mathcal{L}_{\mathrm{OPSD}}^{\mathrm{LLM}}
=
\mathbb{E}_{\hat{o} \sim \pi_\theta(\cdot \mid q)}
\left[
D\!\left(
\pi_{\bar{\theta}}(\cdot \mid q,r),\,
\pi_\theta(\cdot \mid q)
\right)
\right].
\label{eq:llm_opsd}
\end{equation}
This formulation inherits two key properties of on-policy learning: it updates the model on samples produced by the current policy, and the supervision is computed under the same sampled trajectory instead of being borrowed from an external offline distribution.

The challenge in transferring this training paradigm to diffusion models lies in how to construct a stronger context. In LLMs, the extra information $r$ can be natively appended to the input sequence~\cite{chain-of-thought,gpt3}. In diffusion models, however, the desired supervision is an image, and one cannot simply insert the target image into the denoising trajectory in the same way without returning to the standard off-policy SFT setting (e.g, Once the noisy target image is fed directly into the model like traditional training does, the sampling trajectory is disrupted, reducing the process to a supervision paradigm analogous to teacher forcing in large language models~\cite{gpt,gpt2,s2s}.). This challenge suggests that the stronger context in diffusion models needs to be introduced through a representation that enriches the model's conditioning space while leaving the student's roll-outs unchanged. In other words, to make OPSD applicable to diffusion models, we need a mechanism that incorporates target-image information without replacing the student's own sampled states. 

We solve this by utilizing the property of modern diffusion models. As we analyse in Section~\ref{sec:intro} and Figure~\ref{fig:motivation}, current SOTA few-step models often adopt LLM/VLM backbones as their encoders and we find that the subsequent diffusion model can inherit the encoder's in-context capability: when conditioned on the multimodal feature extracted from both the text prompt and the target image, the model can already produce variations that preserve the target concept or style, even without additional training. This observation allows us to instantiate OPSD in diffusion models by treating the target image as in-context supervision, rather than as a direct denoising target.

\begin{figure}[t]
    \centering
    \vspace{-1.5em}
    \includegraphics[width=1\linewidth]{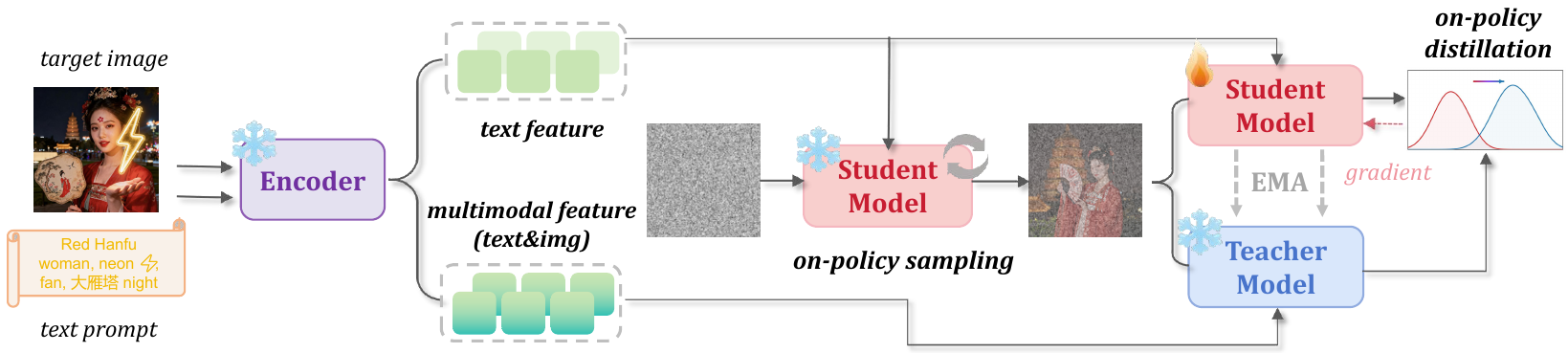} 
    \caption{\textbf{Method overview.} For each training pair, we first pass the prompt alone and the prompt together with the target image through the encoder to obtain $c_s$ and $c_t$, respectively. We then sample a few-step trajectory using the student branch conditioned on $c_s$. After that, the teacher and student predict velocities on the same trajectory states, and the student is updated by Equation~\ref{eq:main_loss}. After training, the teacher branch is discarded, and inference uses exactly the same few-step text-to-image pipeline as the original step-distilled model.}
    \vspace{-1em}
    \label{fig:method}
\end{figure}

\paragraph{Formulating OPSD for diffusion models.}
The overall framework and pseudocode of our method \sname can be seen in Figure~\ref{fig:method} and Algorithm~\ref{alg:method}.
Specifically, we consider the model parameterized by $\theta$, whose velocity field is denoted by $v_\theta(x_t, t, c)$, where $x_t$ is the latent state at time $t \in [0,1]$ and $c$ is the condition feature. During inference, the model defines an ODE trajectory:
\begin{equation}
\frac{d x_t}{d t} = v_\theta(x_t, t, c),
\label{eq:flow_ode}
\end{equation}
which is solved with a small number of time steps by the few-step sampler (e.g, 4 or 8). Let $1 = t_K > t_{K-1} > \cdots > t_0 = 0$ denote the inference schedule. Starting from Gaussian noise $x_{t_K}^{s} \sim \mathcal{N}(0,I)$, the student roll-outs is generated by:
\begin{equation}
x_{t_{k-1}}^{s}
=
\Phi\!\left(
x_{t_k}^{s}, t_k, t_{k-1},
v_\theta(\cdot,\cdot,c_s)
\right),
\quad k=K,\ldots,1,
\label{eq:student_rollout}
\end{equation}
where $\Phi$ denotes the same few-step solver used at test time, and $c_s$ is the student condition.

For each training pair $(x_0, y)$, we construct two conditions from the same encoder:
\begin{equation}
c_s = f_{\mathrm{text}}(y), \qquad
c_t = f_{\mathrm{mm}}(y, x_0),
\label{eq:conditions}
\end{equation}
where $f_{\mathrm{text}}$ encodes only the text prompt and $f_{\mathrm{mm}}$ encodes the multimodal context consisting of the text prompt and the target image. The student is conditioned only on $c_s$, so its inference pathway is exactly the original text-to-image generation process. The teacher is conditioned on $c_t$, which provides additional information about the target concept, style, or preference to be learned.

Given the on-policy trajectory $\{x_{t_k}^{s}\}_{k=1}^{K}$, we evaluate both branches on the same visited states:
\begin{align}
u_k^{s} = v_\theta(x_{t_k}^{s}, t_k, c_s), \quad
u_k^{t} = v_{\bar{\theta}}(x_{t_k}^{s}, t_k, c_t),
\end{align}
where $\bar{\theta}$ denotes the teacher parameters. We then train the student to match the teacher's velocity prediction by minimizing:
\begin{equation}
\mathcal{L}_{\mathrm{\sname}}
=
\mathbb{E}_{(x_0,y)}
\left[
\frac{1}{K}
\sum_{k=1}^{K}
\left\|
u_k^{s} - \mathrm{sg}(u_k^{t})
\right\|_2^2
\right],
\label{eq:main_loss}
\end{equation}
where $\mathrm{sg}(\cdot)$ denotes stop-gradient operation. In this way, the student is optimized on its own roll-outs states, while the teacher provides a stronger supervision signal through multimodal context.

Note that Equation~\ref{eq:main_loss} can be viewed as the diffusion counterpart of Equation~\ref{eq:llm_opsd}. At a high level, the analogy is straightforward: the student's sampled response in LLMs corresponds here to the student's denoising trajectory, and the teacher's stronger prediction under a richer context is realized as a stronger conditional denoising field. The main difference lies in the output space of the model. Autoregressive LLMs produce a discrete token distribution~\cite{gpt,llama}, so the teacher-student alignment can be written directly as a divergence between vocabulary distributions~\cite{opsd,sdft}. Flow-matching diffusion models, by contrast, do not expose such a discrete predictive distribution at each step. Instead, they parameterize the denoising dynamics through a conditional velocity field, whose predictions determine the evolution of the sample trajectory~\cite{flow-matching,rectified-flow,sde}. For this reason, we instantiate the teacher-student alignment in Eq.~\ref{eq:main_loss} as a mean-squared error between velocity predictions on the same on-policy states. Although this objective is not a token-level KL divergence~\cite{kl-d}, it serves the same role in our setting: it pulls the student's conditional generation dynamics toward those of the teacher, thereby aligning the induced trajectory distribution under a stronger multimodal context. The underlying principle therefore remains unchanged: the model learns from its own trajectory under a stronger self-generated supervision signal.

\begin{wrapfigure}{r}{0.62\textwidth}
\vspace{-1em}
\begin{minipage}{0.62\textwidth}
\hrule
\vspace{0.4em}
\captionof{algorithm}{Training Procedure of \sname}
\vspace{-0.2em}
\hrule
\vspace{0.4em}
\begin{algorithmic}[1]
\Require Training pairs $\mathcal{D}=\{(x_0,y)\}$
\Require Inference schedule $\{t_0,\ldots,t_K\}$ and Solver $\Phi$
\Require Base model $v_{\phi}$, Student and Teacher model $v_{\theta}$, $v_{\bar{\theta}}$

\State Initialize student weights $\theta \leftarrow \phi$
\State Initialize teacher weights $\bar{\theta} \leftarrow \phi$
\While{not converged}   
    \State Sample a mini-batch $(x_0,y)\sim\mathcal{D}$
    \State Encode student condition $c_s \gets f_{\mathrm{text}}(y)$
    \State Encode teacher condition $c_t \gets f_{\mathrm{mm}}(y,x_0)$
    \State Initialize $x_{t_K}^{s} \sim \mathcal{N}(0,I)$
     \State Initialize $\mathcal{L}_{\mathrm{\sname}} \gets 0$
    \For{$k = K,\ldots,1$}
    \State $u_k^{s} \gets v_{\theta}(x_{t_k}^{s}, t_k, c_s)$
    \State $u_k^{t} \gets v_{\bar{\theta}}(x_{t_k}^{s}, t_k, c_t)$
    \State $\mathcal{L}_{\mathrm{\sname}} \gets \mathcal{L}_{\mathrm{\sname}} + \frac{1}{K}\|u_k^{s} - \mathrm{sg}(u_k^{t})\|_2^2$
    \If{$k > 1$}
        \State $x_{t_{k-1}}^{s} \gets \mathrm{sg}(\Phi(x_{t_k}^{s}, t_k, t_{k-1}, u_k^{s}))$
    \EndIf
    \EndFor
    \State Update student model $\theta$ by minimizing $\mathcal{L}_{\mathrm{\sname}}$
    \State Update teacher model $\bar{\theta}$ via EMA
\EndWhile
\end{algorithmic}
\label{alg:method}
\vspace{0.4em}
\hrule
\end{minipage}
\vspace{-1em}
\end{wrapfigure}

\paragraph{Discussion on why \sname preserves few-step capability.}
Compared with vanilla SFT, our method does not force the model to fit states induced by the target image that never appear under its own few-step sample trajectory. Instead, both the optimization states and the supervision signal are defined on the student's actual roll-outs, which substantially reduces the mismatch between training and inference. This distinction is particularly important for step-distilled diffusion models, where even small perturbations to the learned few-step dynamics can directly harm generation quality. As a result, \sname provides an on-policy supervised training paradigm for step-distilled diffusion models, enabling them to learn new concepts, styles, or domain preferences from paired image-text data while preserving the original few-step sampling behavior. More discussion and comparison of different training paradigms are provided in Appendix~\ref{appen:method_diff}.

\begin{table*}[t]
\centering
\vspace{-1.5em}
\captionof{table}{\textbf{System-level comparison against baseline methods in LoRA training settings.} The \textbf{best} and \underline{second-best} results on each metric are highlighted in bold and underlined.}
\vspace{-0.5em}
\resizebox{1\linewidth}{!}{
\begin{tabular}{l c c c c c c}
\toprule
{Method}  & DINO-D$\downarrow$ & LPIPS-D$\downarrow$ & VLM-J$\uparrow$ &CLIP-S$\uparrow$ & Quality-S$\uparrow$
& Aesthetic-S$\uparrow$
\\
\arrayrulecolor{black}\midrule

\multicolumn{7}{l}{\textcolor{gray}{\emph{Z-Image-Turbo}}} \\
Base-Model & 0.2373 & 0.7520 & 1.0000 & 0.3043 & 3.5088 & 2.9667\\
Vanilla SFT & 0.2212 & 0.6501 & 1.3293 & 0.3095 & 2.4236 & 2.3582 \\
SFT + LoRA on distilled & 0.1588 & 0.7243 & 1.8525 & 0.3201 & \underline{3.6081} & 3.0027 \\
Dreambooth & 0.0902 & 0.6424 & \underline{3.0625} & \underline{0.3328} & 2.5582 & 2.3755 \\
PSO & \textbf{0.0570} & \textbf{0.4974} & \textbf{3.3333} & 0.2893 & 3.3422 & \underline{3.0820} \\
\rowcolor[RGB]{240,230,245}
\sname\ (ours) & \underline{0.0823} & \underline{0.5803} & \textbf{3.3333} & \textbf{0.3664} & \textbf{3.7965} & \textbf{3.1710} \\

\arrayrulecolor{black!40}\midrule

\multicolumn{6}{l}{\textcolor{gray}{\emph{FLUX.2-klein }}} \\

Base-Model & 0.2386 & 0.7518 & 1.0000 & 0.3020 & 3.5076 & 2.9665\\
Vanilla SFT & 0.2212 & 0.6500 & 1.2083 & 0.3002 & 2.4119 & 2.3595 \\
SFT + LoRA on distilled & 0.1564 & 0.7288 & 1.0000 & 0.3151 & \underline{3.6079} & 3.0021 \\
Dreambooth & 0.1053 & 0.6824 & 3.0283 & \underline{0.3266} & 2.5492 & 2.3683 \\
PSO & \textbf{0.0593} & \textbf{0.5120} & \textbf{3.3015} & 0.2801 & 3.3258 & \underline{3.0566} \\
\rowcolor[RGB]{240,230,245}
\sname\ (ours) & \underline{0.0822} & \underline{0.5648} & \underline{3.1255} & \textbf{0.3689} & \textbf{3.7957} & \textbf{3.1521} \\

\arrayrulecolor{black}\bottomrule
\end{tabular}}
\vspace{-0.5em}
\label{tab:system_compare_lora}
\end{table*}

\section{Experiment}
\subsection{Experimental Setup}
\noindent{\textbf{Implementation.}}
We use Z-Image-Turbo 6B~\cite{z-image} and  FLUX.2-klein 4B~\cite{flux-2} as our baseline model to conduct experiment.  Detailed experimental implementation, including hyperparameter settings, GPU resources, and other training configs, is provided in Appendix~\ref{appen:imp_detail}.

\textbf{Evaluation.}
We use the same inference settings as the original step-distilled model across all methods. We choose to report DINO distance (DINO-D)~\cite{dino}, LPIPS distance (LPIPS-D)~\cite{lpipsscore},  Fr\'echet Inception Distance (FID~\citep{fid}) for testing whether the model can learn from the target images, VLM's judgment of subject or style consistency (VLM-J), CLIP Score (CLIP-S)~\cite{clipscore} for testing whether the model can generalize with the learned new knowledge, the Quality Score (Quality-S) and Aesthetic Score (Aesthetic-S) from the reward model for testing whether the model maintain its few-step sampling capacity, as well as GenEval~\cite{geneval} and DPG~\cite{dpg} score to test whether the model retain its previous knowledge. We also provide a user study for evaluation. Detailed explanation of how the evaluation set is constructed and how each metric is obtained  are in  Appendix~\ref{appen:eval_metric}.

\textbf{Methods for comparison.}
We compare with several representative baseline methods: (a). directly training with vanilla flow-matching loss~\cite{flow-matching} (Vanilla SFT). (b).  training on the original multi-step model then adding LoRA on the step-distilled model (SFT + LoRA on distilled). (c). Dreambooth style training~\cite{dreambooth} (Dreambooth). (d). PSO training~\cite{pso} (PSO). Note that we do not show the results of two-stage training strategies, such as first performing supervised fine-tuning on a multi-step model and then applying step distillation. One practical reason is that,  the exact distillation recipe, including key hyperparameter configurations, is not publicly available for the base model we use. Reproducing such pipelines would therefore require substantial implementation assumptions and could introduce unfairness into the comparison. However, we still provide a comparison with a two-stage variant that uses the open-source DMD method in the Appendix~\ref{appen:compare_twostage} for reference.

\subsection{Main Results}

\begin{figure}[t]
    \centering
    \vspace{-1em}
    \includegraphics[width=0.9\linewidth]{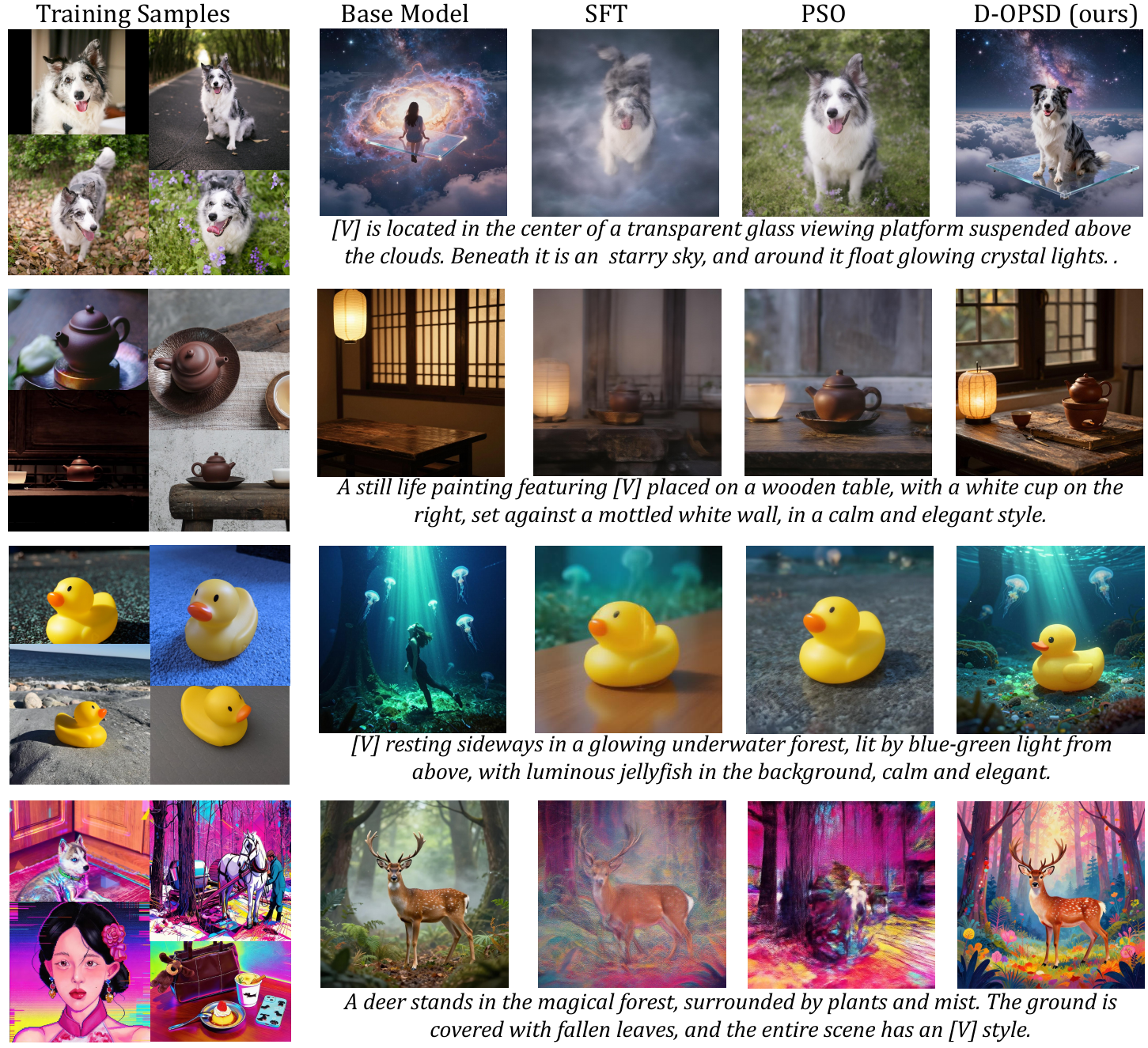}  
    \caption{\textbf{Visual comparison between baseline methods and ours finetuned on Z-Image-Turbo under customized training settings.} Vanilla SFT training sacrifices the original few-step capacity, and PSO suffers from the overfitting to training set, whereas our method enables the step-distilled model to continuously learn new concepts while maintaining the few-step capacity.}
    \vspace{-1em}
    \label{fig:lora_compare}
\end{figure}

\textbf{\sname for LoRA training on small customized dataset.} We first evaluate \sname in the setting of LoRA training on small customized datasets. In this setting, the goal is to learn a new concept from only a few image–text pairs (e.g., 4 examples) while still being able to generalize beyond the training set. We conduct training and evaluation on the DreamBooth dataset~\cite{dreambooth} together with a small amount of stylized data.
As shown in Table~\ref{tab:system_compare_lora}, our method substantially outperforms  SFT style training on DINO-D, LPIPS-D, and VLM-J. Moreover, as illustrated in Figure~\ref{fig:lora_compare}, after training, our method can generalize the newly learned concept beyond the training distribution, e.g., generating the learned object in novel scene  that do not appear in the training data, while preserving the original model’s ability to produce high-quality images with a small number of inference steps. In contrast, other baselines such as SFT and DreamBooth training lose the ability to generate high-quality images under the few-step inference setting, as reflected by the large drops in Quality-S and Aesthetic-S in  Table~\ref{tab:system_compare_lora}, as well as the blurry images shown in the Figure~\ref{fig:lora_compare}. PSO, on the other hand, tends to overfit the training set: although it can capture the target concept, its ability to follow novel instructions degrades substantially, as indicated by the decline in CLIP-S and its failure to generate scenes beyond those in the training data.

\begin{table*}[t]
\centering
\vspace{-1.5em}
\captionof{table}{\textbf{System-level comparison against baseline methods in full-finetuning settings.} The \textbf{best} and \underline{second-best} results on each metric are highlighted in bold and underlined.}
\vspace{-0.5em}
\resizebox{1\linewidth}{!}{
\begin{tabular}{l c c c c c c c}
\toprule
{Method} & FID$\downarrow$ & DINO-D$\downarrow$ & LPIPS-D$\downarrow$  & Quality-S$\uparrow$
& Aesthetic-S$\uparrow$ & GenEval$\uparrow$
& DPG$\uparrow$ 
\\
\arrayrulecolor{black}\midrule

\multicolumn{8}{l}{\textcolor{gray}{\emph{Z-Image-Turbo}}} \\
Base-Model & \underline{48.6858} & \underline{0.1274} & 0.7036 & \underline{3.7624} & \underline{3.5218} & \textbf{0.7543} & \textbf{84.7645}\\
Vanilla SFT  & 82.2036 & 0.1896 & 0.6787 & 2.6121 & 2.4852 & {0.1588} & 69.9746\\
PSO & 88.4343 & 0.1716 & \underline{0.6530} &  2.8653 & 2.6424 & 0.2475 &72.7363\\
\rowcolor[RGB]{240,230,245}
\sname\ (ours) & \textbf{40.4938} & \textbf{0.1088} & \textbf{0.6419} & \textbf{3.8438} & \textbf{3.6195} & \underline{0.7170}  & \underline{84.1116} \\

\arrayrulecolor{black!40}\midrule

\multicolumn{7}{l}{\textcolor{gray}{\emph{FLUX.2-klein }}} \\
Base Model & \underline{45.2335} & \underline{0.1210} & 0.6894 & \underline{3.7626} & \underline{3.4465} & \textbf{0.8155} & \textbf{85.5624} \\
Vanilla SFT  & 98.7893 & 0.2565 & 0.6567 & 2.5038 & 2.4121 & 0.1113 &56.2138 \\
PSO & 82.1438 & 0.2185 & \underline{0.6511} & 2.8066 & 2.5952 & 0.2406 & 63.4174 \\
\rowcolor[RGB]{240,230,245}
\sname\ (ours) & \textbf{38.2932} & \textbf{0.0902} & \textbf{0.6476} & \textbf{3.8328} & \textbf{3.5511} & \underline{0.7298}  & \underline{83.8927} \\

\arrayrulecolor{black}\bottomrule
\end{tabular}}
% \vspace{-0.5em}
\label{tab:system_compare_full}
\end{table*}
\begin{figure}[t]
    \centering
    \vspace{-1.0em}
    \includegraphics[width=1\linewidth]{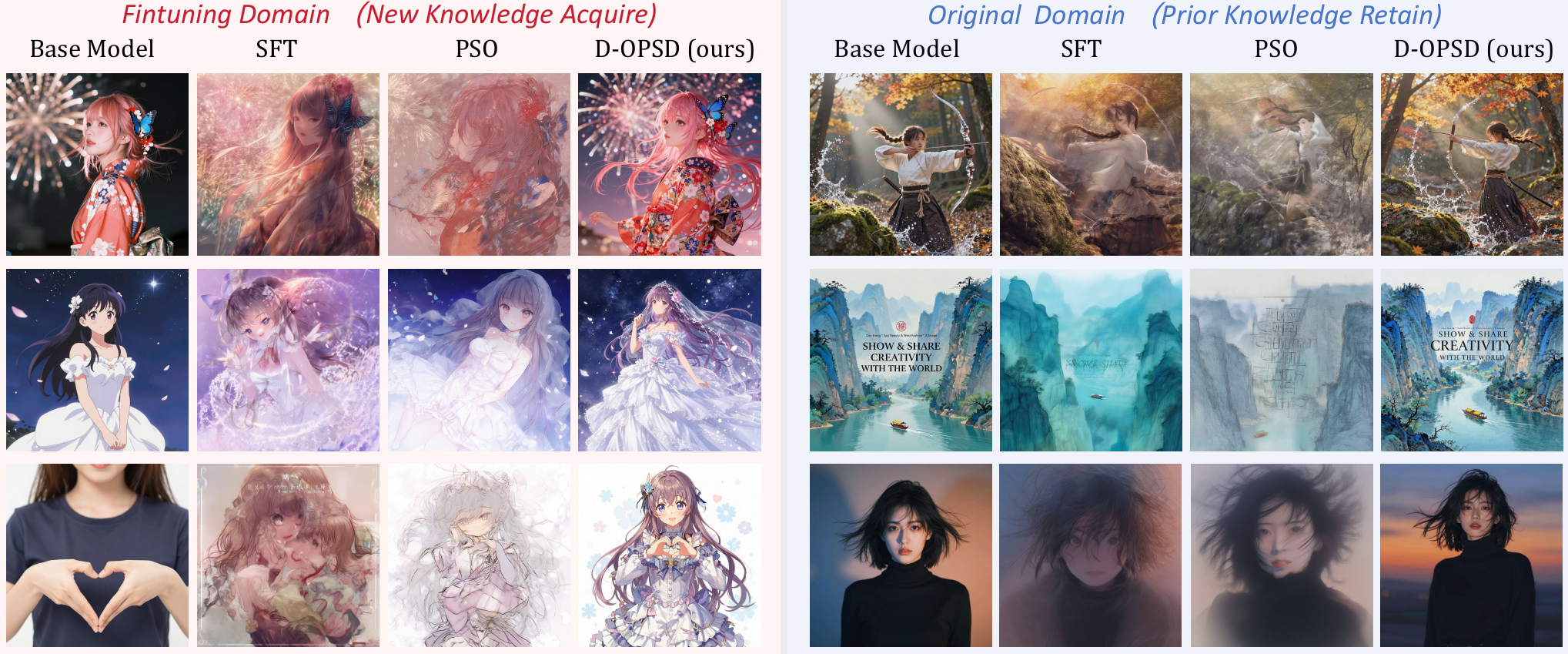} 
    \caption{\textbf{Visual comparison between baseline methods and ours finetuned on Z-Image-Turbo under full-finetuning settings.} SFT and PSO training sacrifices the original few-step capacity, whereas our method enables the step-distilled model to continuously learn to bias target domain while maintaining the few-step capacity as well as the learned knowledge in the original domain.}
    \label{fig:full_compare}
    \vspace{-1.0em}
\end{figure}
\begin{figure*}[t]
    \centering
    \includegraphics[width=1\linewidth]{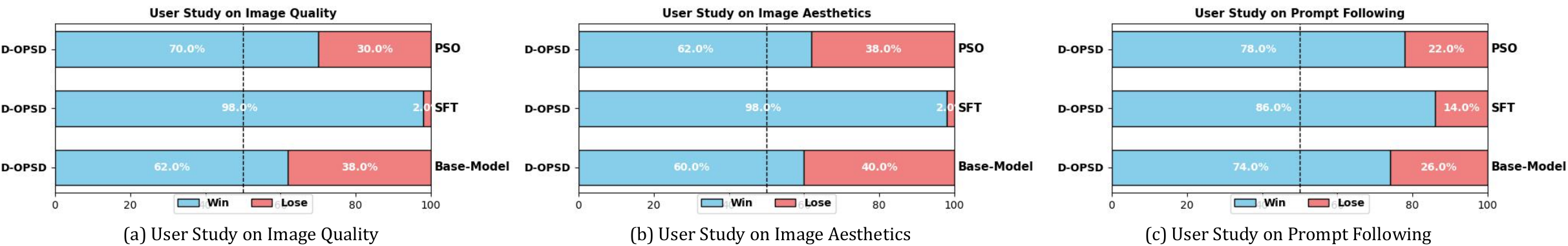}
    \caption{\textbf{User study comparing with baselines.}  All images are generated using identical noise. Our method produces images with superior aesthetics and prompt coherence.}
    \label{fig:user_comp_all}
    \vspace{-1.0em}
\end{figure*}

\textbf{\sname for full finetuning on larger scale dataset.} We next evaluate \sname in the setting of full finetuning on larger scale dataset. In this setting, the goal is to test whether by fine-tuning, the model can be biased towards a certain preference or domain (in our experiment, it is ``anime" domain), and  suffer from catastrophic forgetting of previously learned knowledge. We conduct training and evaluation on the high-quality anime dataset. 
As shown in Table~\ref{tab:system_compare_full}, our method substantially outperforms both the base model and other training methods on FID, DINO-D, and LPIPS-D, suggesting the output of the model after finetuning is more likely to be closer to the target distribution. Meanwhile, our method is still able to adapt to the new distribution while retaining the model’s original knowledge as well as few-step inference ability. This can be observed from the GenEval and DPG results in the Table~\ref{tab:system_compare_full} and Figure~\ref{fig:full_compare}, where our method shows no catastrophic degradation after fine-tuning. Although there is a slight drop in benchmark score compared with the base model, we believe this reflects a trade-off introduced by adapting the model to a new distribution whose domain differs from those emphasized by the benchmarks. In contrast, both SFT and PSO fail to simultaneously adapt to the new domain and preserve the model’s few-step inference capability in the full-finetuning on large-scale dataset setting. This is evident from the sharp declines across multiple metrics in Table~\ref{tab:system_compare_full}, as well as the blurry generated images shown in Figure~\ref{fig:full_compare}.

We also report user study results in Figure~\ref{fig:user_comp_all}.  \sname is preferred over PSO, Vanilla SFT, and even the base model across all three aspects: image quality, aesthetics, and prompt following. The preference margin is particularly large against SFT, indicating that our method better preserves the original few-step generation quality during continual tuning. Notably, \sname also largely outperforms the base model in prompt following, suggesting that the learned target-domain knowledge improves performance without sacrificing overall generation quality.

\subsection{Ablation Study}

\begin{figure}[t]
    \centering
    \includegraphics[width=1\linewidth]{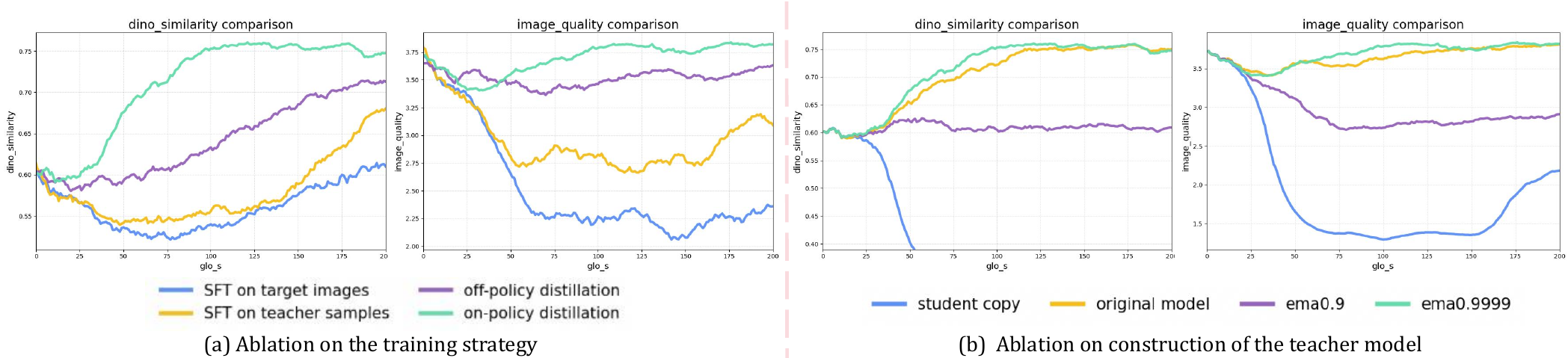} 
    \caption{\textbf{Ablation on (a) the different training strategies, and (b) the different way to build teacher model}. We report the curves across training steps of DINO feature similarity between the generated images and the targets, as well as the Quality Score of the generated images. Training conducted on Z-Image-Turbo with LoRA.}
    \vspace{-1em}
    \label{fig:ablation}
\end{figure}

\textbf{Effect of on-policy self-distillation.} Our method consists of two key components: on-policy sampling and on-policy distillation. To elucidate the role of each component, we conduct four groups of ablation studies in isolation: (1) SFT from target images, which is identical to the vanilla training setting with flow-matching loss.
(2) SFT from teacher samples, where we replace the target images with samples generated by conditioning on multimodal features extracted from the target image and the text prompt, and use these generated samples as the new targets for SFT. (3) Off-policy distillation, where the student model is trained to align with the teacher’s outputs on a fixed dataset. (4) On-policy distillation, which corresponds to our proposed method.
As shown in the two left plots of Figure~\ref{fig:ablation}, vanilla SFT with the flow-matching loss gradually impairs the model’s ability to generate high-quality images with few steps, as reflected by the progressive decline in the Quality Score. Self-distillation-based schemes effectively mitigate this issue. Compared with the off-policy distillation variant, our method achieves the fastest training convergence, as evidenced by the highest DINO similarity to the target images, while simultaneously maintaining the best generation quality.

\textbf{Construction of the teacher model.} 
As our method is a self-distillation framework, we study several way to build the teacher model. First, we find that using the frozen base model as the teacher yields stable training and effective results. We then study the commonly used EMA operations~\cite{ema}, like other self-distillation works~\cite{sra,dino,ibot}, we also find that it requires a large momentum coefficient to stabilize training; for example, directly using the student copy leads to training collapse. In our experiment, using the EMA teacher with the
momentum coefficient 0.9999 leads to the best results, we assume that it is because this can not only extremely smooth the
high-variance alignment target to make training stable, but also tracks the student’s progress for better distillation.

\section{Discussion on Limitations and Future Works}

\textbf{Computation cost.} Like other on-policy distillation method~\cite{opsd,sdft,opd}, our method also need an on-policy roll-outs of student and a teacher inference during training, which results in roughly 4$\times$ computational cost in FLOPs and 2$\times$ training time per iteration compared to vanilla SFT. However, in our task, continually tuning of few-step diffusion models, we consider this cost acceptable. This is because SFT would degrade the model’s few-step generation capability; when the computational cost of the re-distillation stage is taken into account, our method is in fact more resource efficient.

\begin{wrapfigure}{r}{0.5\textwidth}
    \centering
    \vspace{-0.5em}
    \includegraphics[width=\linewidth]{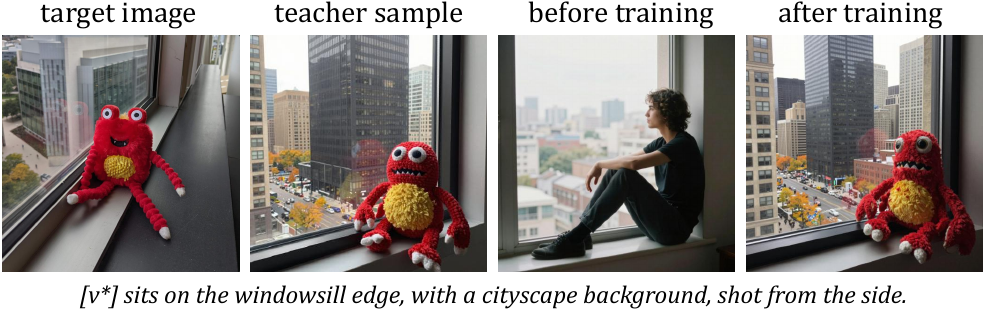}
    \caption{When the teacher model fails to generate images consistent with the concept ID under multimodal condition and therefore cannot provide an effective supervision signal, training will fail.}
    \vspace{-1em}
    \label{fig:failure_case} 
\end{wrapfigure}

\textbf{Requirements for teacher capability.} The success of \sname is contingent upon the base model's in-context  abilities. In specific, as shown in Figure~\ref{fig:failure_case}, even when conditioned on the multimodal feature of target image and the text prompt, the subsequent diffusion model still can not generate the meaningful supervision signals, the training would fail.

\textbf{Future Works.} In this work, we introduce on-policy self-distillation to image generation and show that it is a promising paradigm for continuously training step-distilled diffusion models. Building on this framework, several directions merit further study. First, an important open question is how to construct richer teacher-side context (conditioning). One possibility is to incorporate stronger conditional signals from image editing models~\cite{nanopro,flowedit,flux-kontext} or video generation models~\cite{wan,ltx2,seedance2}. Second, how to leverage other training target in \sname, for example by combining our framework with additional training constraints~\cite{soar,sra}. Third, it is worth exploring whether multi-expert OPD~\cite{deepseekv4,copd} can be introduced into the post-training stage of diffusion models based on \sname loss. A possible strategy is to first train domain-specific experts using RL or SFT, and then distill these experts back into a single base model within our framework. More broadly, We hope our study provide useful insights for future research on post-training and continuous adaptation in diffusion-based generation.

\section{Related Work}
We highlight key related studies here and defer discussion of the others to Appendix~\ref{appen:related}.

\textbf{Step-distilled diffusion models.}  
To accelerate diffusion model inference for enhancing productivity, various timestep distillation methods have been proposed to compress the original model into a generator capable of few-step sampling~\cite{dmd,lcm,add,diff-instruct,kdgen,tdm}. This is typically achieved either by distilling the trajectory~\cite{lcm,hypersd,rcm,piflow} or from the distribution~\cite{dmd,dmd2,dmdr,twinflow,ladd,add,dmdx}. Although a significant effort has been made to ensure the quality of the generated contents while reducing the number of inference steps, continual fine-tuning on these distilled models still faces the challenge of how to preserve their few-step inference capability when learning new things. In this work, we solve this by utilizing the in-context capacity of the diffusion model's encoders to develop an on-policy self-distillation framework that enables the continuous learning of the model under its own supervision, without sacrificing its original few-step inference capability.

\textbf{On-policy self-distillation.}  In the field of large language models,  on-policy distillation is proposed to mitigate the train-test mismatch issues caused by off-policy SFT or knowledge distillation~\cite{opd,survey-opd,opd-tml,Veto}. However, it still requires a stronger external teacher model for guidance. Therefore, on-policy self-distillation is proposed, which enables the model itself to act as a teacher by leveraging its own in-context capabilities within the pre-existing context (e.g, demonstration, answer)~\cite{opsd,opcd,sdft,opsdc,sdrlvr,sd-zero,pid}. Unlike these works, which focused on text generation with autoregressive large language models,  our approach shows how to utilize on-policy self-distillation in image generation for continuously training step-distilled diffusion models.

\section{Conclusion}
In this work, we present \sname, an on-policy self-distillation framework for continually tuning step-distilled diffusion models. Our method builds on the observation that modern diffusion models with LLM/VLM encoders exhibit an emergent in-context capability, enabling the same model to serve as a student under text-only conditioning and as a teacher under stronger multimodal conditioning. By distilling the teacher's predictions along the student's own few-step roll-outs, \sname enables supervised adaptation without external rewards or auxiliary training stages. Experiments on both LoRA adaptation and full fine-tuning show that our method learns new concepts, styles, and domain preferences while preserving few-step generation quality and prior knowledge.

\section{Acknowledgment}

Since the release of Z-Image-Turbo, we are grateful to the community for many interesting explorations into the internal mechanisms of step-distilled models and how to conduct continuous training~\cite{deturbo,latentscaffold, train-zit,reddit-discussion}, and our work is also inspired by these interesting attempts. We cannot list all the works here, but we still want to express our gratitude to all the talented community `artists'!

\clearpage
\appendix
\section*{\centering Appendix}
\section{Investigation of  FLUX.2-klein}
\label{appen:flux_zeroshot}

\begin{figure}[H]
    \centering
    \includegraphics[width=1\linewidth]{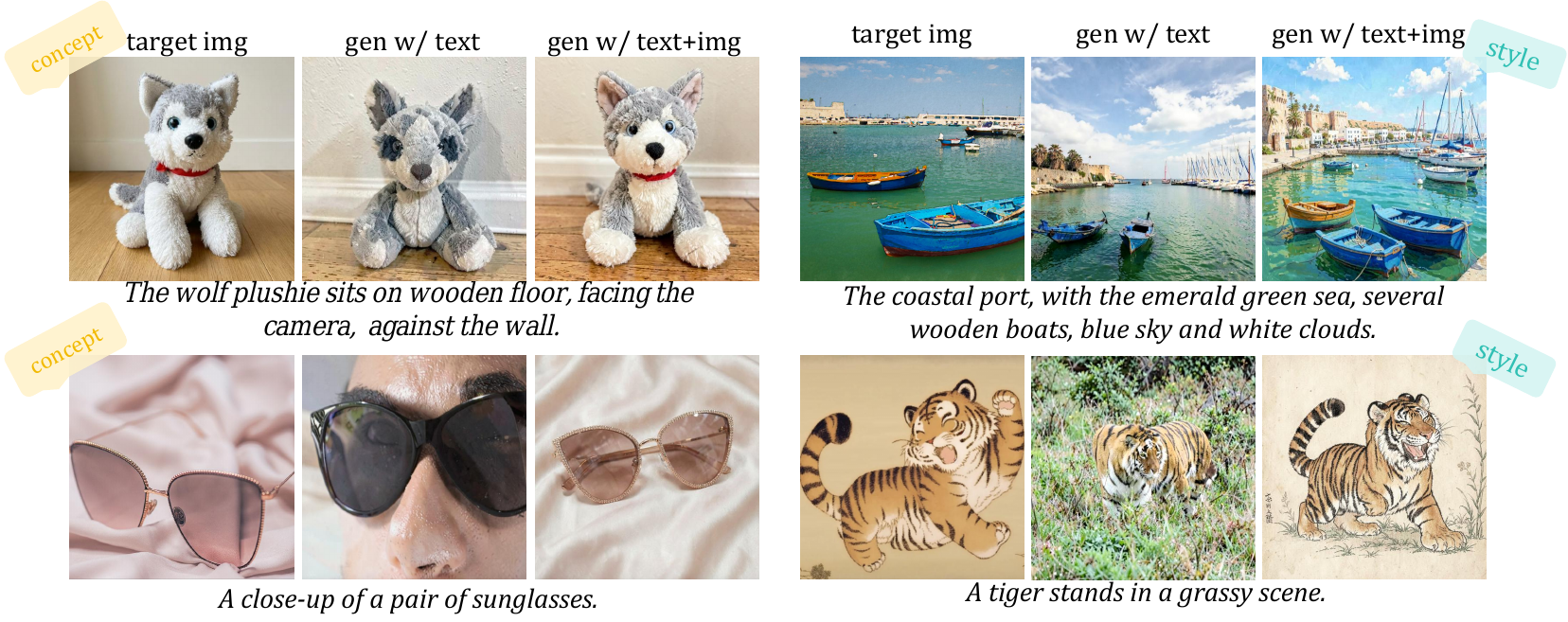}  
    \caption{We also empirically investigate the difference of generated images when conditioned on only the text feature or the multimodal feature of target image and the text prompt using FLUX.2-klein-4B with 4 steps. Similar to Z-Image-Turbo, using multimodal features as condition instead of text-only features allows the model to produce image variations while maintaining the target image’s underlying concept or stylistic identity. }
    \label{fig:motivation_f}
\end{figure}

We also perform a similar analysis with FLUX.2-klein~\cite{flux-2} like those have done in Figure~\ref{fig:motivation}, as shown in Figure~\ref{fig:motivation_f}, the similar behavior can be seen, suggesting that this inherited in-context capability is broadly applicable to diffusion models that employ LLM/VLMs as encoders.

\section{Discussion and Comparison of Different Training Paradigms}
\label{appen:method_diff}
Table~\ref{tab:comapre_method} summarizes the main differences among representative training paradigms for continually tuning step-distilled diffusion models. We compare them along four dimensions: the form of supervision signal, whether training is on-policy, whether an additional reward model or reward function is required, and whether the training process matches the model's inference behavior.

\begin{table*}[h]
\centering
\captionof{table}{Comparison of different training paradigms regarding the source of the supervision signal, the learning paradigm (whether its in an on-policy way), the necessity of an auxiliary reward model, and the consistency between training-time inference-time state.}
\vspace{-0.2em}
\resizebox{1\linewidth}{!}{
\begin{tabular}{l c  c c c}
\toprule
{Method}  & Supervision Signal & On-Policy &  Reward Model &Train-Inference Match\\
\\
\arrayrulecolor{black}\midrule
{SFT} & GT velocity & $\times$ &$\times$ & $\times$\\
{RL-offline (e.g, Diffusion-DPO~\cite{diffusiondpo}, PSO~\cite{pso})} & GT velocity pairs $(+/-)$ &$\times$ &$\times$ & $\times$
\\
{RL-online (e.g, ReFL~\cite{refl}, flow-grpo~\cite{flowgrpo}} & reward function  & $\checkmark$  & $\checkmark$ & $\checkmark$\\
\rowcolor[RGB]{240,230,245}
{\sname (ours)} & self-distilled velocity & $\checkmark$ &$\times$ & $\checkmark$
 \\
\arrayrulecolor{black}\bottomrule
\end{tabular}}
\vspace{-0.5em}
\label{tab:comapre_method}
\end{table*}

\paragraph{Vanilla SFT.}
Vanilla supervised fine-tuning optimizes the model using the target image as the source of supervision. In flow-matching models, this means that training is performed on noised states constructed from the ground-truth image, with the model regressed toward the corresponding ground-truth velocity. While this objective is standard for training diffusion models from scratch, it is mismatched to continual tuning of step-distilled models. The reason is that both the optimization states and the supervision signal are induced by the target image, rather than by the model's own few-step sampling trajectory. As a result, although the model can absorb new concepts or styles from the training pairs, we suppose that it may do so by altering the distilled generation dynamics that are responsible for high-quality few-step sampling.

\paragraph{Offline RL-style methods.}
A natural alternative is to replace direct regression on a single target with preference-style or pairwise supervision. Representative examples include Diffusion-DPO~\cite{diffusiondpo} and PSO~\cite{pso}\footnote{PSO can be regarded as a variant of the Diffusion-DPO~\cite{diffusiondpo} for the step-distilled model because it only conducts training at the few-step sampling timestep, but it still uses the target image state as input and uses the ground truth velocity for supervision..}. These methods can be viewed as offline RL-style objectives, in the sense that their supervision is still derived from a fixed dataset (use target image related states as inputs and rely on ground-truth velocity-based pairwise supervision). Therefore, although methods like PSO try to specially adapt the few-step models, their optimization states and supervision signal are still not fully induced by the student's own current distribution. This also explains why, in our experimental results, PSO can often learn the target appearance but tends to overfit the training set in the small dataset and fails to learn in the large-scale settings.

\paragraph{Online RL-style methods.}
Online RL methods, such as ReFL~\cite{refl} and flow-GRPO~\cite{flowgrpo}, are conceptually more suitable for preserving the behavior of step-distilled models because they optimize the model on its own sampled trajectories. In these methods, the model first generates images on-policy, and the resulting samples are then scored by a reward function or reward model. As such, both the optimization states and the supervision signal are tied to the current sampling process, substantially reducing the mismatch between training and inference. This is also an important reason why some studies that perform RL on the step-distilled model can make the model align with human preference without compromising the original few-step ability~\cite{dmdr,tdmr1}. However, this advantage comes at the cost of requiring a well-designed reward function or preference model. In practical customization scenarios, especially when secondary developers only possess a small number of image-text pairs, such reward design is often the main bottleneck.

\paragraph{Our method.}
Our method occupies a different point in this design space. Like online RL, \sname is on-policy: optimization is performed on the student's own few-step roll-outs, so the model is always updated on states that it actually visits at inference time. More importantly, the supervision signal is also defined on these same states. Instead of introducing the target image as an external denoising target, \sname uses it only to enrich the teacher's condition through multimodal in-context encoding, and supervises the student with self-distilled velocity predictions evaluated on the student's current trajectory. At the same time, unlike RL-based approaches, \sname does not require any external reward model or manually designed reward function. In this sense, \sname combines the main advantage of online optimization, train-inference consistency, with the practicality of supervised learning from paired image-text data.

\paragraph{Why this distinction matters for step-distilled models.}
For multi-step diffusion models, moderate train-test mismatch can sometimes be tolerated because iterative denoising provides room for error correction~\cite{sde,ddpm,ddim,flow-matching}. Step-distilled models are less forgiving: with only a few denoising steps, even small deviations in the learned dynamics can directly harm image quality~\cite{dmd2,selfforcing}. For this reason, in our setting, it is important not only whether the training states are aligned with the model's own sampling trajectory, but also whether the supervision signal is defined under that same trajectory. This is the main motivation behind the design in Table~\ref{tab:comapre_method}. Among the compared paradigms, \sname is the only one that simultaneously satisfies four desirable properties for continual tuning of few-step models: it is on-policy, does not require a reward model, preserves train-inference consistency, and still incorporates target image-text pairs into training through self-distillation.

\section{Comparison with Two-stage Training Pipeline}
\label{appen:compare_twostage}
\begin{table*}[t]
\centering
\vspace{-1.5em}
\captionof{table}{\textbf{System-level comparison with two-stage training pipelines.} We compare one-stage D-OPSD with SFT on the few-step model, SFT on the multi-step model, and a two-stage pipeline that applies DMD distillation after multi-step SFT. Experiments are conducted with Z-Image and Z-Image-Turbo. The \textbf{best} and \underline{second-best} results on each metric are highlighted.}
\vspace{-0.5em}
\resizebox{1\linewidth}{!}{
\begin{tabular}{l c c c c c}
\toprule
Method & DINO-D$\downarrow$ & Quality-S$\uparrow$ & GenEval$\uparrow$ & Inference NFE$\downarrow$ & Training GPU hours$\downarrow$ \\
\arrayrulecolor{black}\midrule
\multicolumn{6}{l}{\textcolor{gray}{\emph{LoRA customization setting}}} \\
SFT on Z-Image-Turbo & 0.2212 & 2.4236 & - & \textbf{8} & \textbf{5} \\
SFT on Z-Image & \textbf{0.0547} & \underline{3.3488} & - & 100 & \textbf{5} \\
SFT on Z-Image + DMD & 0.1024 & 3.3372 & - & \textbf{8} & 5 + 882 \\
\textbf{\sname (ours)} & \underline{0.0823} & \textbf{3.7965} & - & \textbf{8} & \underline{9} \\

\arrayrulecolor{black!40}\midrule
\multicolumn{6}{l}{\textcolor{gray}{\emph{Full-finetuning setting}}} \\
SFT on Z-Image-Turbo & 0.2565 &  2.5038 & 0.1588 & \textbf{8} & \textbf{1066} \\
SFT on Z-Image & \textbf{0.0843} & \underline{3.3801} & \underline{0.6578} & 100 & \textbf{1066} \\
SFT on Z-Image + DMD & 0.1094 & 3.3652 & 0.6084 & \textbf{8} & 1066 + 2054 \\
\textbf{\sname (ours)} & \underline{0.1088} & \textbf{3.8438} & \textbf{0.7170} & \textbf{8} & \underline{1918} \\
\arrayrulecolor{black}\bottomrule
\end{tabular}}
\vspace{-1em}
\label{tab:system_compare_two_stage}
\end{table*}

In this section, we compare D-OPSD with a two-stage training pipeline that first fine-tunes a multi-step model on the target data and then distills the tuned model back into a few-step model. We conduct the comparison using open-source Z-Image as the multi-step model and Z-Image-Turbo as its few-step counterpart. For the distillation stage, we adopt the open-source DMD algorithm~\cite{dmd,dmd2}. In the LoRA customization setting, we first train a LoRA adapter on Z-Image, and then train an additional step-distillation LoRA on the tuned model (Main model with LoRA). In the full-finetuning setting, both the tuning stage and the DMD distillation stage are performed with full model updates. The Training GPU hours are tested with the H800 GPU.

The results are reported in Table~\ref{tab:system_compare_two_stage}. Directly applying SFT to Z-Image-Turbo is computationally simple, but it substantially weakens the model's few-step generation capability, especially in the LoRA setting, where the Quality-S score drops from the original few-step model quality regime. This is consistent with our main observation: vanilla SFT optimizes on target-image-induced states rather than on the states visited by the model's own few-step sampler, thereby perturbing the distilled few-step dynamics. Fine-tuning the multi-step model instead gives the best DINO-D in both settings, indicating strong fitting to the target domain. However, this model requires 100 NFEs at inference time and therefore loses the main efficiency advantage of step-distilled models. Moreover, in the full-finetuning setting, it shows a clear degradation in Quality-S and GenEval, suggesting that adapting to the target domain alone does not preserve the generation behavior desired from the few-step model.

The two-stage pipeline partially addresses the inference cost by applying DMD after multi-step SFT, reducing inference to 8 NFEs. However, it introduces substantial extra training cost and still does not recover the overall performance of the tuned model. In particular, the DMD-distilled variants obtain lower Quality-S and GenEval scores than D-OPSD, indicating degradation in both few-step generation quality and retained prior knowledge. This limitation is partly practical: the strongest distillation recipes used by SOTA few-step foundation models are often not fully public, including key hyperparameters and engineering details~\cite{ddmd,dmdr}. Therefore, reproducing the closed-source distillation pipeline of the original base model is not feasible for most secondary developers. With the publicly available DMD implementation, the additional distillation stage is expensive and can still introduce performance loss.

In contrast, D-OPSD performs continual tuning in a single stage. It learns the new target-domain information while keeping both optimization and supervision on the student's own roll-outs, which preserves the original few-step sampling behavior. Although D-OPSD requires on-policy student roll-outs and teacher inference during training, leading to roughly $4\times$ FLOPs and about $2\times$ wall-clock training time per iteration compared with vanilla SFT, we consider this cost acceptable for continual tuning of few-step diffusion models. The roll-out and teacher-inference computations do not require gradient computation and can be further optimized in engineering. More importantly, once the cost of re-distillation is included, D-OPSD is more resource-efficient than the two-stage pipeline while achieving better few-step quality and stronger retention of prior knowledge.

\section{Implementation Details.}
\label{appen:imp_detail}

\begin{wrapfigure}{r}{0.6\textwidth}
    \centering
    % \vspace{-1.em}
    \includegraphics[width=\linewidth]{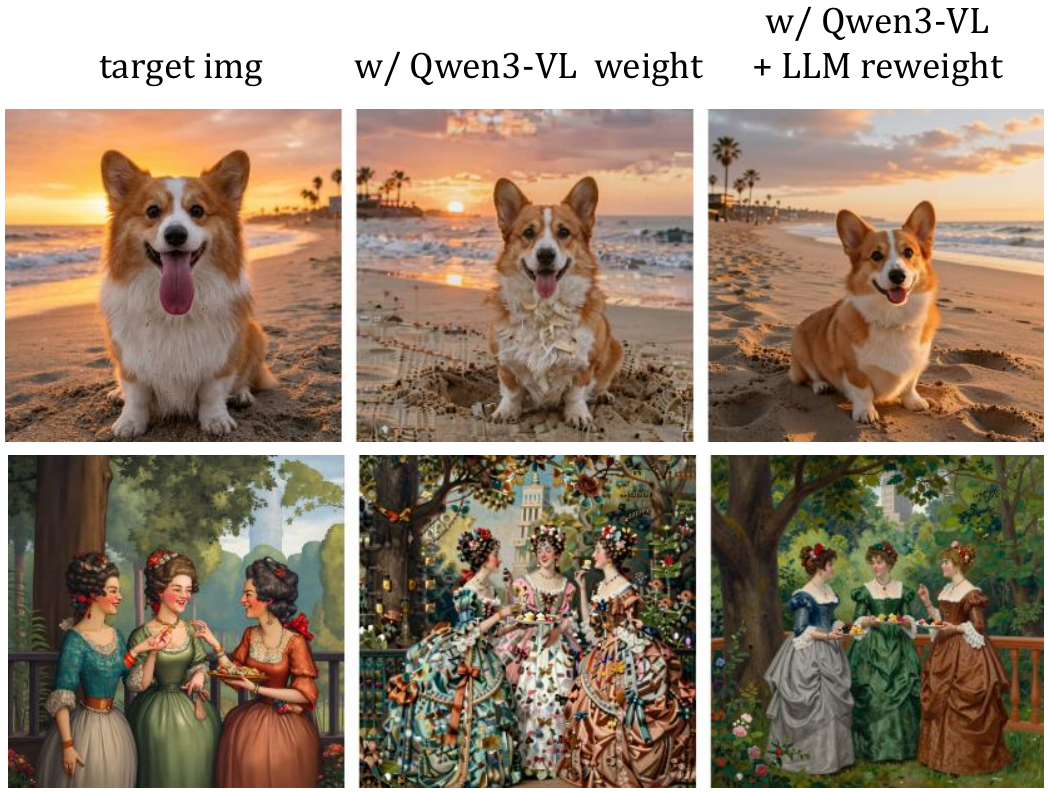} 
    \caption{Comparison of generated images of Z-Image-Turbo conditioned on multimodal feature using Qwen3-VL 4B and  Qwen3-VL 4B with LLM part reweighted by Qwen3-4B LM. }
    \vspace{-1em}
    \label{fig:reweight} 
\end{wrapfigure}

We now provide a detailed training and implementation settings of \sname as follows:

\textbf{Encoder settings of student and teacher.} During training, for the student model, we use the original text encoder of the diffusion model (for both Z-Image-Turbo and  FLUX.2-klein, text prompts are encoded using Qwen3-4B~\cite{qwen3}). For the teacher model, since target image information must be incorporated, a straightforward solution is to replace Qwen3-4B with the corresponding Qwen3-VL-4B~\cite{qwen3vl}. However, in practice, we find that this naive substitution introduces high-frequency artifacts and excessive sharpening in the generated images (See Figure~\ref{fig:reweight}, the middle column). We attribute this issue to a mismatch of the feature space between inference and training distribution.\footnote{Since Qwen3-VL is a continually trained variant of Qwen3-LM, the model can still drive image generation, but its output distribution no longer aligns well with that of the diffusion model’s original training setup.}

To address this issue, we replace the weights of the LLM component in Qwen3-VL-4B with those from the more compatible Qwen3-4B, while keeping the ViT and Connector weights unchanged. In this way, we preserve multimodal in-context capability while making the output distribution as consistent as possible with that seen during diffusion model training (See Figure~\ref{fig:reweight} the last column). Notably, this operation can be viewed as approximately reverting the VLM to the stage where the Connector has been trained, but the LLM parameters remain unchanged~\cite{llava,qwen2vl,internvl}. Although the final VLM exhibits stronger multimodal capabilities, the model at this earlier stage still retains a certain degree of multimodal understanding. This reflects a trade-off between preserving the output image quality of the final diffusion model and maintaining the in-context capability provided by the VLM. As more LLMs/VLMs evolve toward native multimodal architectures~\cite{gemini2.5,gemini3,qwen3.5}, we expect this trade-off to be naturally alleviated in future diffusion models using stronger multimodal encoders, such as Qwen3.5~\cite{qwen3.5}.

\textbf{LoRA training on small customized dataset}:
\vspace{-0.07in}
\begin{itemize}[leftmargin=0.3in]
		\item \textbf{Z-Image-Turbo}: We use LoRA with rank 64 and alpha 128 to finetune the model. We set the total batch size to 4 and the learning rate to 4e-5\footnote{The learning rate for Z-Image-Turbo is consistently set higher than that for FLUX.2-klein because we find that the norm of the parameters in Z-Image-Turbo is greater than that of other models, including FLUX.2-klein. Consequently, a proportionally larger learning rate is required to maintain effective parameter updates during optimization.}. The momentum coefficient of EMA decay is set to 0.9999 by default. (Note that since the model is trained with LoRA, we only need to use one main weight and then perform the EMA operation on the LoRA weights. This can effectively save the memory). The model is trained with 1K iterations on a single H800 GPU.
        \item \textbf{FLUX.2-klein}: We also use LoRA with rank 64 and alpha 128 to finetune the model. We set the total batch size to 4 and the learning rate to 1e-5. The momentum coefficient of EMA decay is set to 0.9999 by default. The model is trained with 1K iterations on a single H800 GPU.
\end{itemize}

\textbf{Full finetuning on larger scale dataset}:
\vspace{-0.07in}
\begin{itemize}[leftmargin=0.3in]	
		\item \textbf{Z-Image-Turbo}: In this setup, we unlock all the parameters in the diffusion transformer backbone. We set the total batch size to 256 and the learning rate to 3e-5. The momentum coefficient of EMA decay is set to 0.9999 by default. The model is trained with 10k iterations on 32 H800 GPUs.
        \item \textbf{FLUX.2-klein}: We also unlock all the parameters in the diffusion transformer backbone. We set the total batch size to 256 and the learning rate to 8e-6. The momentum coefficient of EMA decay is set to 0.9999 by default. The model is trained with 10K iterations on 32 H800 GPUs.
\end{itemize}

\section{More Details of Evaluation}
\label{appen:eval_metric}
We now provide the detailed evaluation protocols.

\textbf{LoRA training on a small customized dataset.} In this setting, we attempt to follow the community’s common secondary fine-tuning setup, where a concept is learned through LoRA training on the base model using only a small number of text–image pairs (e.g, less then 10). Thus, we adopt DreamBooth-style data~\cite{dreambooth} for both fine-tuning and evaluation. To ensure stable results, the scores in Table~\ref{tab:system_compare_lora} are averaged over multiple categories (30 concept class and 10 style class) rather than selected from one or two cherry-picked cases.  We use the following metric to evaluate the model:
\vspace{-0.07in}
\begin{itemize}[leftmargin=0.3in]	

\item \textbf{DINO distance (DINO-D)}: We first take the training captions of the training images and use the LLM to paraphrase the captions without changing their core semantics. These rewritten prompts are then used to generate images with the fine-tuned model. We compute the cosine distance between the DINO features of the generated images and those of the corresponding target images. For a specific model, we use DINOv3-ViT-S-plus ~\cite{dinov3} as the feature extractor.

\item \textbf{LPIPS distance (LPIPS-D)}: Similar to DINO-D, we compute the LPIPS distance~\cite{lpipsscore} between the generated images and the corresponding target images. For a specific model, we adopt the commonly used VGG network~\cite{vgg} for this metric.

\item \textbf{VLM’s judgment of subject or style consistency (VLM-J)}: We fix the learned training concept and ask the LLM to construct four groups of prompts that differ from the training prompts but still contain the same concept. For object concepts, the new prompts vary aspects such as scene and composition; for style concepts, the new prompts describe specific image contents different from those seen during training. We use these prompts to generate images with the fine-tuned model, and then feed the generated images together with the target image into the VLM, which is asked to assess the similarity of the concept and assign a score followed the rule of scoring: 4 points(basically the same); 3 points(relatively similar); 2 points (slightly similar); 1 point (completely different). For specific model, we use Qwen3-VL-8B-Instruct~\cite{qwen3vl} for evaluation.

\item \textbf{CLIP Score (CLIP-S)}: Using the same images as in VLM-J, we further evaluate whether the model still follows the non-concept parts of the prompt, such as the background. Specifically, we compute the image-text alignment between the generated image and the prompt using CLIP~\cite{clip,clipscore}. Note that we replace the special DreamBooth class token ([V]) with the original class name when computing this metric. For a specific model, we use DFN-CLIP-H~\cite{dfnclip} for evaluation.

\item \textbf{Quality Score (Quality-S) and Aesthetic Score (Aesthetic-S)}: We use our internal VLM-based reward model for scoring. Compared with traditional open-source CLIP-based reward models, such as ImageReward~\cite{refl} and PickScore~\cite{pickscore}, the model does not require the text prompt used for image generation as input, and provides more reliable scores thanks to large-scale training with dedicated human preference annotations and a reasoning process before giving a final score during inference. Note that we do not explicitly optimize either of these two scores during fine-tuning. Moreover, all compared methods are evaluated under the same scoring protocol and criteria. 
\end{itemize}

\textbf{Full finetuning on larger scale dataset}: In this setting, we fully fine-tune the model like a normal large-scale SFT. Given that we use the latest state-of-the-art open-source models~\cite{z-image,flux-2} as our baseline, most publicly available open-source datasets are unsuitable for our full finetuning experiment, as their overall quality is lower than that of the data used to train these baseline models. Therefore, we rely on an in-house dataset of 25K high-quality anime images. We use the following metric to evaluate the model: 

\vspace{-0.07in}
\begin{itemize}[leftmargin=0.3in]
\item \textbf{Fréchet Inception Distance (FID)~\cite{fid}}: We randomly sample 2K data from the training set, and use the fine-tuned model to generate the images from these prompts. the two sets of images are then fed into the Inception-v3 network~\cite{inception-model} to extract features. Assumes both feature distributions are multivariate Gaussian and computes the Fr´echet distance between them.

\item \textbf{DINO distance (DINO-D)}: The images generated for calculating FID score will also be used to calculate DINO distance, following the rules used in the above LoRA evaluation settings.

\item \textbf{LPIPS distance (LPIPS-D)}: Similar to DINO-D, the images generated for calculating FID score  will also be used to calculate LPIPS distance following the rules used in the above LoRA evaluation settings.

\item \textbf{Quality Score (Quality-S) and Aesthetic Score (Aesthetic-S)}: Similar to both DINO-D and LPIPS-D, the images generated for calculating FID score  will also be used to calculate Quality Score and Aesthetic Score from the Reward model we introduced above.

\item \textbf{GenEval and DPG score}: We follow the evaluation settings
in these benchmarks~\cite{geneval,dpg} to generate images and calculate score. Note that we use prompts from the original benchmark instead of the Prompt-Enhanced (PE)~\cite{pe} variants for generating images.

\end{itemize}

\textbf{User study.} Following prior work~\cite{dmd2,tdm,dreambooth}, we conduct a comprehensive human evaluation to assess overall generation performance. Specifically, we sample 50 prompts from the small-scale LoRA customization setting and another 50 prompts from the large-scale full-finetuning setting, yielding 100 evaluation prompts in total. For the small-scale setting, we additionally provide annotators with the corresponding reference images or style exemplars, so that prompt following can be judged with respect to both the textual instruction and the target concept/style to be learned. To ensure fair and systematic assessment, we compare paired outputs from our method and the baselines under randomized presentation, and ask annotators to evaluate them along three dimensions: image quality, aesthetic quality, and prompt following. We then aggregate the preferences from both settings into a single overall score, and report the final results in Figure~\ref{fig:user_comp_all}.

\section{More Related Works}
\label{appen:related}
We now provide a detailed literature review of other related work.

\textbf{Image generation diffusion models.}  
Diffusion models have achieved remarkable success in image generation. Early works mainly study unconditional or class-conditional generation, demonstrating that progressively denoising noise can produce highly realistic images~\cite{adm,edm}. This framework was then extended to text-to-image synthesis, where text conditions are injected into the denoising network to guide image generation according to natural language descriptions~\cite{dalle,ldm}. Later, latent-space diffusion models and stronger text-conditioning designs further improve both efficiency and generation quality, making diffusion models the mainstream solution for text-to-image generation~\cite{sdxl,pixartalpha,ldm}. Building on this, recent studies introduce more scalable backbones and objectives, such as diffusion transformers and flow/rectified-flow formulations, which continue to push the frontier of image fidelity, prompt following, and training scalability~\cite{dit,sit,pixartalpha,sd3,flux-1,sana}. Meanwhile, unlike earlier models that usually rely on CLIP~\cite{clip} or T5~\cite{t5} as the condition encoder, the latest high-performance image generation models increasingly adopt large language models or vision-language models as the encoder~\cite{luminaimage,z-image,qwenimage,flux-2}. Our method is built upon this evolution: we show that the diffusion model can benefit from the in-context capability inherited from these modern encoders, which makes on-policy self-distillation feasible for continuously tuning step-distilled diffusion models.

\textbf{Knowledge distillation for diffusion model.} 
Besides step-distillation for faster sampling, other forms of knowledge distillation~\cite{kd,kd-survey} is also widely used in diffusion models. Commonly, a more powerful pretrained model is often used to guide the diffusion model during training~\citep{flux1-lite,repa,tinyfusion,reg}. Meanwhile, self-distillation frameworks also demonstrates effectiveness even without external components (e.g, stronger model)~\cite{sra,sra2,sddit,selfflow,elt}. For example, SRA~\cite{sra} aligns the output latent representation of the diffusion transformer in earlier layer with higher noise to that in later layer with lower noise to progressively enhance the overall representation learning during only generative process and accelerate the training convergence of the model, which has also been demonstrated in subsequent study~\cite{selfflow} to be applicable in multiple modalities (video, audio, etc.). Our work is also related to self-distillation, while our approach is conducted in an on-policy way for preserving few-step inference capacity during supervised fine-tuning.

\textbf{Diffusion model fine-tuning.}  
A large body of work studies how to adapt pretrained diffusion models to new concepts, styles, or downstream domains~\cite{dreambooth,Ip-adapter,textual-inversion,custom-diffusion,controlnet,refvton}. Representative approaches include standard supervised fine-tuning with all model parameters, subject-driven customization methods such as DreamBooth~\cite{dreambooth}, textual inversion~\cite{textual-inversion}, and parameter-efficient adaptation techniques such as LoRA~\cite{lora}. Among them, full-parameter SFT remains a common choice when sufficient paired data are available, as it directly optimizes the full generative model on the target distribution. However, such standard fine-tuning paradigms are mainly developed for conventional multi-step diffusion models, and they do not explicitly consider whether the adapted model can preserve the few-step inference capability of a step-distilled generator. In particular, directly applying the commonly used denoising or flow-matching objective during SFT would lead to degradation of their original fast-sampling behavior. In contrast, our work focuses on continuously tuning already distilled diffusion models and specifically targets preserving their native few-step generation ability during supervised adaptation.

% \clearpage
\bibliographystyle{splncs04}
\bibliography{main}

\end{document}